\documentclass[final,3p,times,twocolumn]{elsarticle}

\usepackage{amssymb}
\usepackage{amsmath}

\usepackage{adjustbox}
\usepackage{comment}
\usepackage{multicol}
\usepackage{multirow}
\usepackage{xcolor}
\usepackage[colorlinks=true,
            linkcolor=purple,
            urlcolor=purple!80,
            citecolor=cyan,
            anchorcolor=black]{hyperref}

\myfooter[C]{\href{https://doi.org/10.1016/j.bspc.2025.108611}{\color{black}{Publication doi: https://doi.org/10.1016/j.bspc.2025.108611}}}

\AtBeginDocument{%
  \hypersetup{
    colorlinks=true,
    linkcolor=purple,
    urlcolor=purple!80,
    citecolor=cyan,
    anchorcolor=black
  }%
}

\begin{document}

\begin{frontmatter}



\title{An Efficient Dual-Line Decoder Network with Multi-Scale Convolutional Attention for Multi-organ Segmentation}


\author[label1,label2]{Riad Hassan}
\ead{riad@cse.green.edu.bd}
\author[label1]{M. Rubaiyat Hossain Mondal}
\ead{rubaiyat97@iict.buet.ac.bd}
\author[label3]{Sheikh Iqbal Ahamed}
\ead{sheikh.ahamed@marquette.edu}
\author[label4]{Fahad Mostafa}
\ead{fahad.mostafa@asu.edu}
\author[label5]{Md Mostafijur Rahman}
\ead{mostafijur.rahman@utexas.edu}


\affiliation[label1]{organization={Institute of Information and Communication Technology, Bangladesh University of Engineering and Technology},
            state={Dhaka},
            country={Bangladesh}}

\affiliation[label2]{organization={Computer Science and Engineering, Green university of Bangladesh},
            state={Dhaka},
            country={Bangladesh}}

\affiliation[label3]{organization={Computer Science, Marquette University},
            city={Milwaukee},
            state={Wisconsin},
            country={USA}}

\affiliation[label4]{organization={New College of Interdisciplinary Arts and Sciences, Arizona State University},
            city={Tempe},
            state={Arizona},
            country={USA}}
            
\affiliation[label5]{organization={Electrical and Computer Engineering, The University of Texas at Austin},
            city={Austin},
            state={Texas},
            country={USA}}

\begin{abstract}
Proper segmentation of organs-at-risk is important for radiation therapy, surgical planning, and diagnostic decision-making in medical image analysis. While deep learning-based segmentation architectures have made significant progress, they often fail to balance segmentation accuracy with computational efficiency. Most of the current state-of-the-art (SOTA) methods either prioritize performance at the cost of high computational complexity or compromise accuracy for efficiency. This paper addresses this gap by introducing an efficient dual-line decoder segmentation network (EDLDNet). In addition to noise-free decoder, the proposed method features a noisy decoder, which learns to incorporate structured perturbation at training time for better model robustness, yet at inference time only the noise-free decoder is executed, leading to lower computational cost. Multi-Scale Convolutional Attention Modules (MSCAMs), Attention Gates (AGs), and Up-Convolution Blocks (UCBs) are further utilized to optimize feature representation and boost segmentation performance. By leveraging multi-scale segmentation masks from both decoders, a mutation-based loss function is also utilized to enhance the model's generalization. The proposed method outperforms SOTA segmentation architectures on four publicly available medical imaging datasets (Synapse, ACDC, SegThor, and LCTSC). EDLDNet achieves SOTA performance with an 84.00\% Dice score on the Synapse dataset, surpassing baseline model like UNet by 13.89\% in Dice score while significantly reducing Multiply-Accumulate Operations (MACs) by 89.7\%. Compared to recent approaches like EMCAD, the proposed EDLDNet not only achieves higher Dice score but also maintains comparable computational efficiency. The outstanding performance across diverse datasets establishes EDLDNet's outstanding generalization, computational and robustness efficiency. The source code, pre-processed data, and pretrained weights are available at \href{https://github.com/riadhassan/EDLDNet}{https://github.com/riadhassan/EDLDNet}.
\end{abstract}







\end{frontmatter}




\section{Introduction}
\label{sec:intro}

\begin{figure}[!htb]
\includegraphics[width=\linewidth, height=6.75cm]{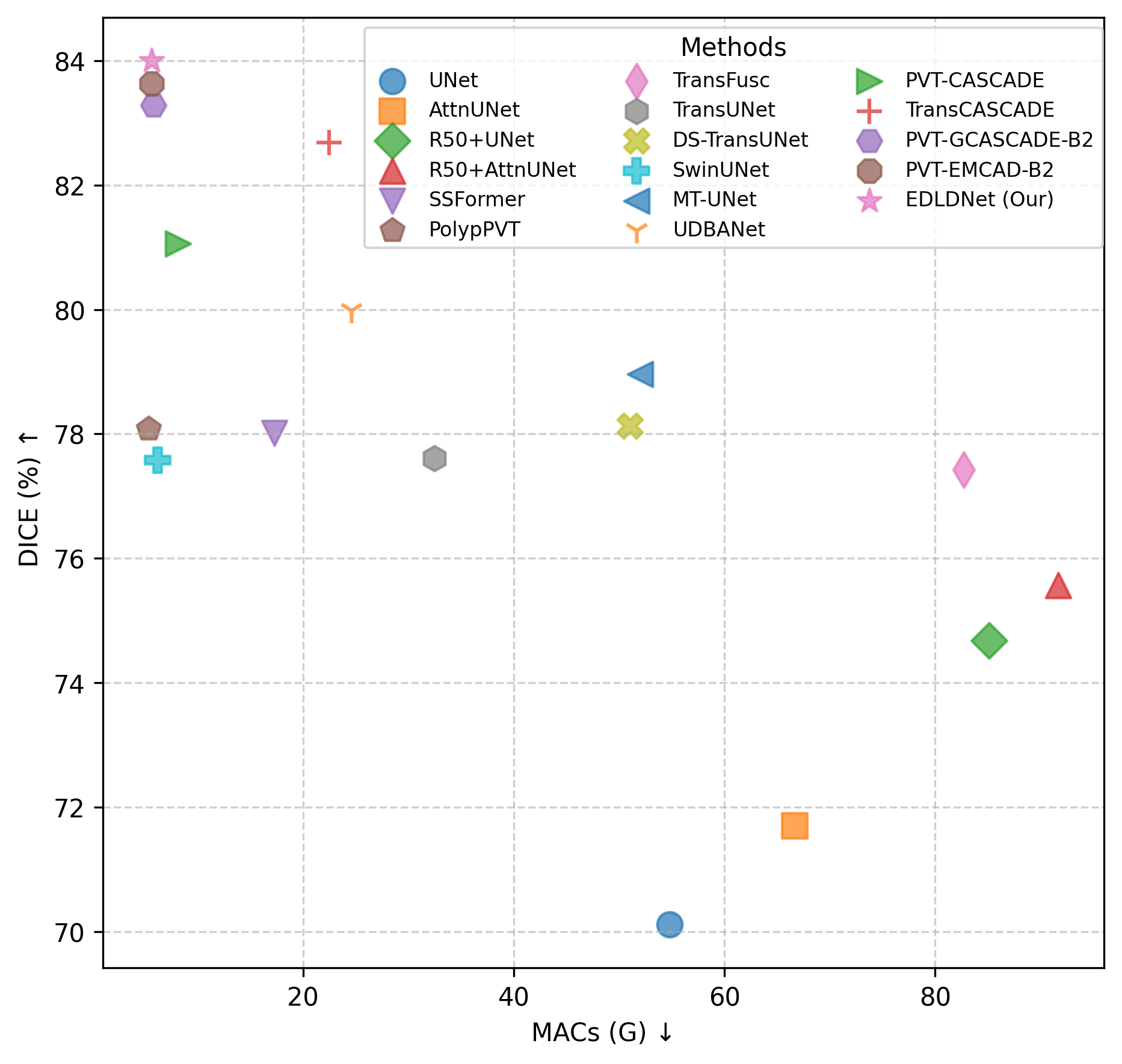}

\caption{Dice score vs MACs count for different segmentation methods over synapse dataset. The figure presents that the proposed method offers the highest segmentation dice score with minimum MACs count.} 
\label{fig:mac_vs_dice}
\end{figure}

Accurate delineation of organs-at-risk (OAR) is critical in radiation therapy, computer-aided surgery, and treatment planning, where precision can significantly affect patient's outcomes \cite{FU2021107,Nijkamp2022}. However, the identification and segmentation of OARs remain challenging and time-consuming, often involving manual processes that are prone to errors and potentially harmful consequences \cite{SHAO2024107955}. Recent advancements in deep learning (DL) have led to notable improvements in OAR segmentation. Different U-shaped networks, UNet \cite{Ronneberger2015}, UNet++ \cite{zhou2020unet}, nnUNet \cite{Isensee2021} have become the standard for medical image segmentation. Attention mechanisms \cite{Ozan2018,Woo_2018_ECCV} have been incorporated into these models to enhance feature representation and improve pixel-level classification accuracy. Uncertainty aware segmentation networks extract uncertainty information to generate attention \cite{10635587} and refine segmentation masks \cite{10.1371/journal.pone.0304771}. While those models have demonstrated better performance, they still encounter major challenges because of high computational demands. Recently, vision transformers \cite{dosovitskiy2021an} have shown significant potential in medical image segmentation tasks by utilizing self-attention mechanisms to capture long-range dependencies. To enhance segmentation performance further, hierarchical vision transformers such as SSFormer \cite{10.1007/978-3-031-16437-8_11}, TransUNet \cite{chen2021transunet}, DS-TransUNet \cite{ds-transunet}, TransFuse \cite{transFUse}, SwinUNet \cite{cao2021swinunet}, PolypPVT \cite{dong2021polyp}, MT-UNet \cite{wu2021mtunet}, MISSFormer \cite{huang2021missformer}, TransCASCADE \cite{Rahman_2023_WACV}, PVT-GCASCADE \cite{Rahman_2023_WACV}, and EMCAD \cite{Rahman_2024_CVPR} have been developed. These methods are high computation and resource intensive. This restricts their practical application, particularly where computational resources are often limited and real-time precise performance is essential.

To overcome these limitations, we propose a segmentation network featuring a dual-line decoder with an efficient multi-scale convolutional attention architecture. This method introduces a unique training strategy: one decoder operates with imposed noise termed as \textit{noisy decoder} while the other remains \textit{noise-free}. During training, both decoders generate segmentation maps in multiple stages that are mutated to create a comprehensive set of combinatorial masks which are used for regularizing the network to increase robustness. In testing or deployment scenarios, only the noise-free decoder is utilized which provides accurate and reliable segmentation with reduced computational overhead. This architecture achieves superior segmentation performance by demonstrating 13.89\% improvement over baseline methods, while 89.7\% more computationally efficient (see Figure \ref{fig:mac_vs_dice}).

The main contributions of this paper are summarized as follows:
\begin{enumerate}
    \item Introduce \textit{noisy} and \textit{noise free} dual-line decoders, both of which are used in model training to produce multi-scale robust segmentation masks for improved regularization of the model, while only the \textit{noise free} decoder is utilized during inference which preserves the computationally efficiency.
    
    \item Implement a mutative implicit logit ensemble technique utilizing multi-level logits derived from both \textit{noisy} and \textit{noise-free} decoders during training. The regularization with mutative logits enhances robustness and noise adaptability in segmentation tasks.
    
    \item Conduct extensive experiments on four publicly available multi-organ segmentation datasets which show the proposed method outperform state-of-the-art (SOTA) methods. It improves 13.89\% in terms of segmentation accuracy (Dice) and 89.7\% computation reduction with compared to the baseline. Compared to recent methods, proposed method delivers a higher Dice score while maintaining comparable computational efficiency.
\end{enumerate}

The remainder of the paper is structured as follows: Section \ref{sec:relt_work} highlights on CNN-based and Transformer-based related work; then, Section \ref{sec:method} presents Methodology. Experimental details, results of experiments on four publicly available datasets, and ablation studies are discussed in Section \ref{sec:exp and result}. Finally, the overall conclusion is drawn in Section \ref{sec:Conclusion}.

\section{Related Work}
\label{sec:relt_work}
Recent advancements in deep learning have led to the development of diverse models leveraging convolutional neural networks (CNNs) and transformers architecture where each addressing specific challenges of segmentation accuracy and computational efficiency.

\subsection{CNN Based Approaches}
CNNs have historically been the backbone of medical image segmentation due to their ability to learn spatial features hierarchically. U-Net \cite{Ronneberger2015} revolutionized the field with its encoder-decoder structure and skip connections. Building on U-Net, several advancements introduced novel components to address specific challenges and enhance segmentation performance. Attention UNet \cite{Ozan2018} integrated attention gates to focus on relevant regions and suppress irrelevant background information which improved segmentation accuracy. R2UNet \cite{8556686} incorporated recurrent residual connections to capture temporal dependencies and refine feature representation iteratively. UNet++ \cite{zhou2020unet} utilized nested dense skip pathways to reduce semantic gaps between encoder and decoder feature maps which enhanced feature integration across scales. nnUNet \cite{Isensee2021} introduced automatic configuration of hyper-parameters to adapt the model to dataset-specific characteristics that eliminated manual tuning and improving generalization. MultiResUNet \cite{IBTEHAZ202074} improved feature representation by incorporating residual connections, enabling more efficient gradient flow and better convergence during training. Uncertainty-aware segmentation networks, such as UDBANet \cite{10635587}, extract uncertainty information to generate attention maps that guide the model to focus on areas of high confidence, while UDBRNet \cite{10.1371/journal.pone.0304771} refines segmentation masks by leveraging uncertainty to reduce errors in boundary predictions. Though these CNN based network have achieved remarkable success, transformer based methods outperform due to its capacity to capture global relationships.

\subsection{Transformer Based Models}
Transformer-based models have revolutionized the field of image segmentation by leveraging their ability to capture global relationships between image pixels. Originally intended for natural language processing (NLP) tasks \cite{vaswani2017attention, devlin2018bert}, transformers have attained superior outcomes in computer vision when modified for image analysis \cite{dosovitskiy2021an, touvron2021training}. The Vision Transformer (ViT) \cite{dosovitskiy2021an} incorporated transformer encoders into computer vision and achieved competitive outcomes through pre-training on extensive datasets like ImageNet21K. Similarly, DeiT \cite{touvron2021training} enabled effective training on limited datasets through techniques such as knowledge distillation. Furthermore, TransUNet \cite{chen2021transunet} integrates CNNs for local feature extraction with transformers for global attention. Additionally, Swin-Unet \cite{cao2021swinunet} employs Swin Transformer blocks \cite{Liu_2021_ICCV} in a U-shaped architecture to improve the encoding and decoding phases. Meanwhile, SSFormer \cite{10.1007/978-3-031-16437-8_11} employs a self-supervised transformer framework to leverae global attention for robust feature extraction. Moreover, MT-UNet \cite{wu2021mtunet} integrates multi-scale transformers into the U-Net architecture, combining the strengths of local and global contextual modeling. PolypPVT \cite{dong2023PolypPVT} uses pyramid vision transformers (PVT) \cite{Wang2022} as foundational architectures and integrates CBAM \cite{Woo_2018_ECCV} modules to enhance decoding efficacy. Building on this, PVT-CASCADE, TransCASCADE \cite{Rahman_2023_WACV}, CSSNet \cite{SHAO2024107955}, G-CASCADE \cite{rahman2024g}, and EMCAD \cite{Rahman_2024_CVPR} introduced attention modules to progressively refine features during decoding which addressed challenges in hierarchical feature aggregation and ensuring more accurate delineation of complex structures. In those approach, improvements in segmentation accuracy typically come at the cost of increased computational requirements due to the higher complexity of self-attention mechanisms and model depth. We propose a segmentation network that enhance segmentation performance without adding computational demands through an optimized architecture.

\begin{figure*}[!htb]
\includegraphics[width=\linewidth, trim={0.5cm 2.8cm 0cm 0cm}, clip]{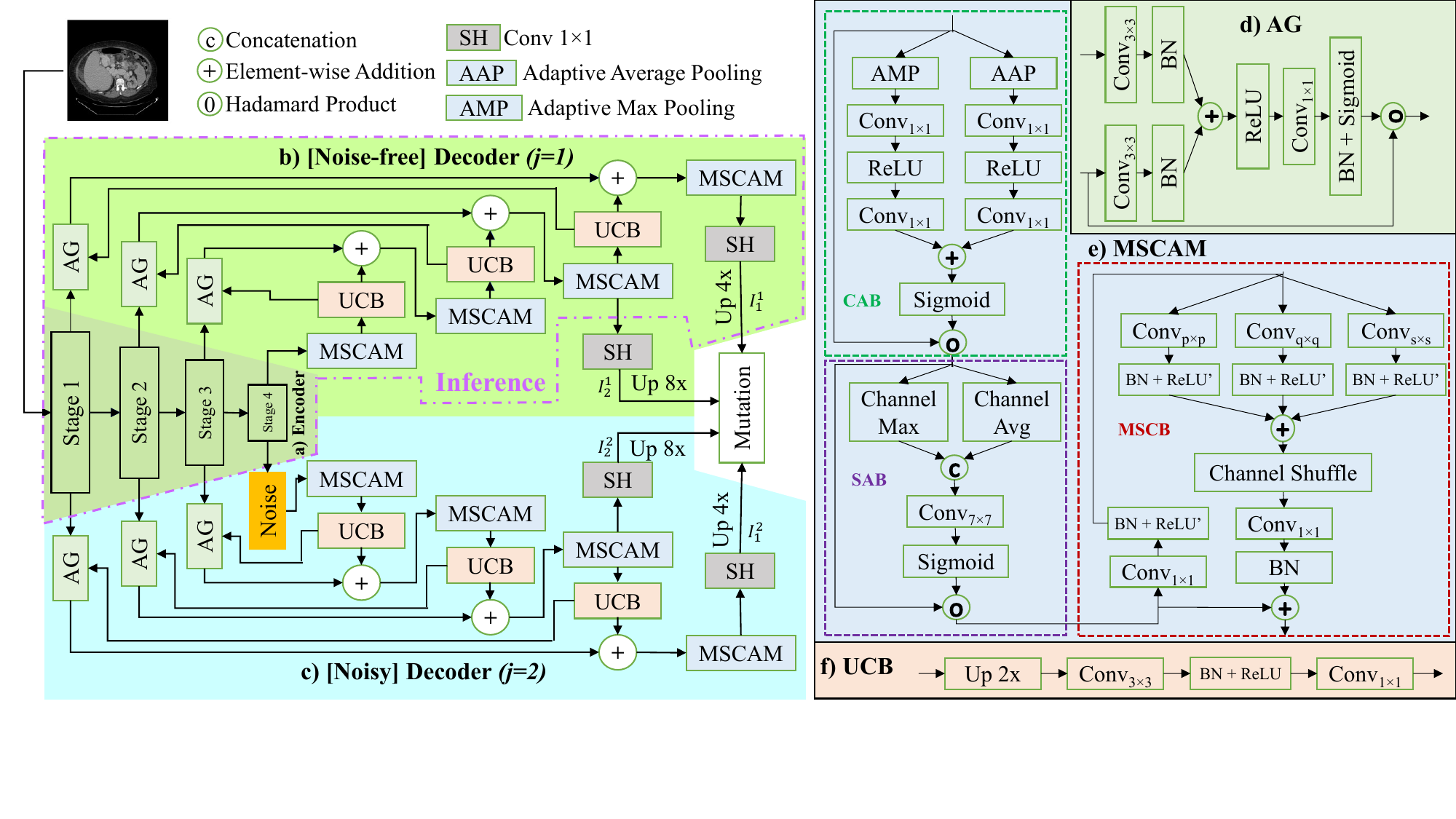}

\caption{The overall architecture of proposed method. a) Pyramid Vision Transformer (PVT) based four stage encoder. b) Noise free decoder comprised of attention gate (AG), Up-convolution block (UCB), Multi-stage Cascaded Attention Module (MSCAM) and Segmentation Head (SH). c) Noisy Decoder is identical with noise free decoder except an uniformly distributed random noise layer in between encoder and decoder (only used during training, not in inference). d) Attention Gates (AGs) to refine the feature maps by integrating skip connections through a gated attention mechanism. e) Multi-scale Convolutional Attention Module (MSCAM) to enhance feature maps. f) Up-Convolution Block (UCB) for up-sampling and further feature enhancement.} 
\label{fig:architecture}
\end{figure*}

\section{Methodology}
\label{sec:method}
In this proposed segmentation network, we employs a vision transformer-based encoder from PVTv2, coupled with dual parallel decoders—a noisy decoder and a noise-free decoder. Each decoder integrates Multi-scale Convolutional Attention Modules (MSCAMs), Up-Convolution Blocks (UCBs), and Attention Gates (AGs) to enhance feature representation and segmentation accuracy. Unlike the noise-free decoder, the noisy decoder incorporates an additional noise layer in its architecture. During training, both decoders generate multiple segmentation logits for mutation-based regularization, while only the noise-free decoder is active during inference to ensure computational efficiency. The complete architecture of the proposed method is illustrated in Figure \ref{fig:architecture}. The subsequent sections detail the encoder and decoder designs, the mutation strategy for training optimization, and the process for generating the final segmentation mask during inference.

\subsection{Encoder}
The proposed architecture employs the PVTv2-B2 transformer as the encoder, chosen for its efficient hierarchical feature extraction capabilities and less computation requirement which are essential for efficient medical image segmentation \cite{Wang2022}. The encoder initiates processing through an overlapping patch embedding layer using a $4 \times 4$ convolutional kernel with stride 4. The four-stage hierarchical architecture is configured with embedding dimensions of $[64, 128, 320, 512]$, utilizing $[1, 2, 5, 8]$ attention heads and $[3, 4, 6, 3]$ transformer blocks per stage, respectively. Each stage incorporates spatial-reduction attention (SRA) with reduction ratios of $[8, 4, 2, 1]$. The MLP layers employ expansion ratios of $[8, 8, 4, 4]$, with Layer Normalization ($\epsilon = 1 \times 10^{-6}$) and learnable QKV biases. Stochastic depth regularization (drop path rate = 0.1) is applied while disabling standard dropout (rate = 0.0) for optimal feature learning. This configuration generates multiscale features labeled as $E_1$, $E_2$, $E_3$ and $E_4$ from four stage of the encoder. These features are then fed into decoders, with $E_4$ integrated into the upsampling path and $E_1$, $E_2$ and $E_3$ utilized within the skip connections.

\subsection{Decoder}
A dual-line decoder with multi-scale convolutional attention is developed to process the extracted features in vision transformer based encoder. Though the dual parallel decoding line are used only in training phase, only one decoder is used in testing or real life applications. The decoders contains MSCAMs to enhance feature maps, UCBs for up-sampling and further feature enhancement, and AGs to refine the feature maps by integrating skip connections through a gated attention mechanism which are described in Section \ref{sec:MSCAM}, and \ref{sec:ucb}, and \ref{sec:AM} respectively. Both decoders $(j= 1, 2)$ are identical except a perturbation layer $\mathcal{N}$ is imposed between the encoder and second decoder ($j=2$) presented in Equation \ref{eqn:noise}. Inspired from \cite{10635587}, the perturbation layer consist of uniformly distributed random noise with distribution $U(-0.3, 0.3)$ which is presented in Equation \ref{eqn:N}. The architecture of the noise free and noisy decoder are presented in Figure \ref{fig:architecture} (b) and Figure \ref{fig:architecture} (c) respectively.

Specifically, four MSCAMs are employed to process the pyramid features $E_1$, $E_2$, $E_3$, and $E_4$ extracted from four stages of the encoder. The refined feature maps from each MSCAM are then upscaled using UCBs and combined with outputs from the corresponding AGs which are presented in Equations \ref{eqn:attention} and \ref{eqn:dec}. 
\begin{equation}
    \mathcal{N}(x) = x + \epsilon, \quad \epsilon \sim U(-0.3, 0.3)
    \label{eqn:N}
\end{equation}
\begin{eqnarray}
    \label{eqn:noise}
        D_4^{j} =
          \begin{cases}
              MSCAM(E_4),  &j = 1 \\
              MSCAM(\mathcal{N}(E_4)),  &j = 2\\
          \end{cases}
\end{eqnarray}
\begin{equation}
    \label{eqn:attention}
    atn_i^j = AG(E_{i}, UCB(D_{i+1}^j)) \quad i = 3, 2, 1; j=1,2
\end{equation}
\begin{equation}
    \label{eqn:dec}
    D_i^{j} =MSCAM(UCB(D_{i+1}^j) \oplus atn_i^j),  i = 3, 2, 1; j=1,2
\end{equation}
Where, $E_i$ represents $i^{th}$ stage encoded features. $D_i^j$ and $atn_i^j$ represents decoded features and attention of $j^{th}$ $(j= 1, 2)$ decoder line and $i^{th}$ stage respectively. 

As the upper stage of decoder carry significant information of segmentation, last two stage decoded features are considered from both decoders (i.e. $D_i^j$ where, $i = 1,2 $ and $ j= 1,2$) followed by $1 \times 1$ convolution and bilinear up scaling ($4x$ and $8x$ for $i=1$ and $i=2$ respectively) to match the size of segmentation ground truth mask represented in Equation \ref{eqn:upscale}. 
\begin{eqnarray}
I_i^j =
\begin{cases}
UpScale_4(Conv_{1\times 1}(D_i^j)),  &i = 1, j = 1, 2 \\
UpScale_8(Conv_{1\times 1}(D_i^j)),  &i = 2, j = 1, 2 \\
\end{cases}
\label{eqn:upscale}
\end{eqnarray}

During training, the segmentation all masks ($I$) presented in Equation \ref{eqn:outputset} are used for mutation ( See Section \ref{sec:mutation}) followed by loss calculation and back propagation. During testing or inference, only $I_{1}^{1}$ is considered as the main segmentation mask which is compared with ground truth for evaluation.
\begin{equation}
\label{eqn:outputset}
    I = \{I_1^1, I_2^1, I_1^2, I_2^2\}
\end{equation}

\subsubsection{Multi-scale Convolutional Attention Module}
\label{sec:MSCAM}
To enhance feature refinement, the Multi-scale Convolutional Attention Module (MSCAM) is employed into the proposed network like \cite{Rahman_2024_CVPR}. It comprised with channel attention, spatial attention, and multi-scale depth-wise convolutions (MSDC) blocks to refine feature maps effectively while maintaining computational efficiency. The MSCAM can be expressed as Equation \ref{eqn:MSCAM}. The architecture of the MSCAM is illustrated in Figure \ref{fig:architecture} (e).
\begin{equation}
    \text{MSCAM}(x) = \text{MSCB}(\text{SAB}(\text{CAB}(x)))
    \label{eqn:MSCAM}
\end{equation}
where $x$ is the input feature map, and $\text{CAB}(\cdot)$, $\text{SAB}(\cdot)$, and $\text{MSCB}(\cdot)$ denote the Channel Attention Block, Spatial Attention Block, and Multi-scale Convolution Block, respectively.

\textbf{CAB:} Channel attention is utilized to prioritize the importance of individual channels, highlighting the most relevant features while downplaying less significant ones. Essentially, the CAB determines which feature maps should receive greater emphasis. CAB can be expressed as Equations \ref{eqn:socab} and \ref{eqn:CAB}.
\begin{equation}
    S_{cab}(\cdot) = W_2(ReLU(W_1(\cdot)))
    \label{eqn:socab}
\end{equation}
\begin{equation}
    \text{CAB}(x) = \sigma(S_{cab}(\text{P}_\text{mx}(x)) \oplus S_{cab}(\text{P}_\text{av}(x))) \otimes x
    \label{eqn:CAB}
\end{equation}

Here, $\text{P}_\text{mx}(x)$ and $\text{P}_\text{av}(x)$ represent the maximum and average pooling operations, respectively. $W_1$ and $W_2$ are point-wise convolutional layers, $ReLU(\cdot)$ \cite{nair2010rectified} is the activation function, $\sigma(\cdot)$ denotes the Sigmoid function, and $\otimes$ signifies element-wise multiplication. $S_{cab}(\cdot)$ represent the sequential operation of $W_1$, $ReLU$, and $W_2$.

\textbf{SAB:} Spatial attention is employed to replicate the way the human brain focuses on particular regions of an input image. In essence, the SAB identifies the areas of importance within a feature map and amplifies those features. This mechanism improves the model's capability to detect and respond to key spatial details. The SAB can be presented as Equation \ref{eqn:sab}.
\begin{equation}
    \text{SAB}(x) = \sigma\left(\text{Conv}\left(\text{Concat}(\text{P}_\text{cmx}(x), \text{P}_\text{cav}(x))\right)\right) \otimes x
    \label{eqn:sab}
\end{equation}
Here, $\text{P}_\text{cmx}(\cdot)$ and $\text{P}_\text{cav}(\cdot)$ perform channel-wise maximum and average pooling, respectively. $\text{Conv}(\cdot)$ denotes a convolution with a large kernel size ($7 \times 7$), and $\text{Concat}(\cdot, \cdot)$ represents channel-wise concatenation. The symbol $\otimes$ indicates element-wise multiplication.

\textbf{MSCB:} Multi-Scale Convolution Block is utilized to improve feature representation in the cascaded expanding path. Inspired by the inverted residual block (IRB) architecture introduced in MobileNetV2 \cite{Sandler_2018_CVPR}, the MSCB incorporates a unique approach by employing depth-wise convolutions at varying scales. Additionally, it leverages channel shuffle techniques \cite{8578814} to facilitate channel mixing across different groups, thereby enhancing the expressiveness and efficiency of the feature extraction process. First the input features are passed through sequentially ($1 \times 1$) point wise convolution, batch normalization which is presented in Equation \ref{eqn:smscb} as $S_{mscb}$. After activation with $ReLU$ activation layer, the output features is pass to the three parallel multi scale convolution with karnel size $1\times1$, $3\times3$, and $5\times5$ followed by batch normalization \cite{ioffe2015batch} and modified relu $ReLU'$ \cite{krizhevsky2010convolutional} as described in Equation \ref{eqn:mrelu}. Then the features from three parallel lines are element wise added which is presented in equation \ref{eqn:msc}. Then the output from previous layer is sent to the channel shuffle ($\text{Ch}_\text{sh}$) technique \cite{8578814}. Finally the the features are passed through the sequential layer $S_{mscb}$. The MSCB can be represented by equation \ref{eqn:mscb}.
\begin{equation}
    S_{mscb}(\cdot) = Conv_{1\times 1}(BN(\cdot))
    \label{eqn:smscb}
\end{equation}
\begin{equation}
    ReLU'(\cdot) = Min(Max(\cdot, 0), 6)
    \label{eqn:mrelu}
\end{equation}
\begin{equation}
    \text{MSC}(\cdot) = \sum_{k \in {1,2,3}} ReLU'(BN(Conv_{k\times k}(\cdot)))
    \label{eqn:msc}
\end{equation}
\begin{equation}
    \text{MSCB}(\cdot) =  S_{mscb}(ReLU'(\text{Ch}_\text{sh}(\text{MSC}(S_{mscb}(\cdot)))))
    \label{eqn:mscb}
\end{equation}

\subsubsection{Up Convolution Block}
\label{sec:ucb}
The number of channel reduces in every stage and reversely, during decoding the number of channel increases. The input feature $x_{ch}$ is passed through up convolution for increasing the number of channel where the sequence of channel number $ch \epsilon \{512,320,128,64\}$. The channel number follows the the sequence of  Then sequential $3\times3$ depth wise convolution, batch normalization and $ReLU$ operations are performed to enhance the features. Finally, ($1 \times 1$) convolution is performed to reduce the channel to fit this output to the next stage input. The process is presented in Equation \ref{eqn:ucb} and the architecture is presented in Figure \ref{fig:architecture} (f).
\begin{equation}
    \text{UCB}(\cdot) = Conv_{1\times1}(ReLU(BN(Conv_{3\times3}(UpConv(\cdot)))))
    \label{eqn:ucb}
\end{equation}

\subsubsection{Attention Gate}
\label{sec:AM}
Attention gate is implemented in every stage of the decoder to aggregate feature maps where relevant features are activated and suppress the irrelevant features. This attention gate takes two input where one from up-convolution block of the previous stage of decoder and another from corresponding stage skip connection of encoder which are termed as $x_u$ and $x_s$ respectively. Unlike ($1 \times 1$) convolution in Attention-UNet \cite{Ozan2018}, $3 \times 3$ grouped convolution \cite{Rahman_2024_CVPR} is employed for both input $Conv_{3\times3}(x_u)$ and $Conv_{3\times3}(x_s)$ followed by batch normalization operation is performed which is presented in equation \ref{eqn:GC}. Then element wise addition operation is perform to combine the both features. After that the features are passed through sequential $ReLU$, $1 \times 1$ convolution, batch normalization and $Sigmoid$ activation layers which are presented in Equation \ref{eqn:So}. Output from the sequential operation is element wise multiplied with $x_u$ to get the final output from attention module $x_{atten}$. The entire process is expressed as following equation \ref{eqn:AM}. The architecture of the attention gate is presented in Figure \ref{fig:architecture} (d).
\begin{equation}
    GC(\cdot) = BN(Conv_{3\times3}(\cdot))
    \label{eqn:GC}
\end{equation}
\begin{equation}
    SO_{ag}(\cdot) = Sigmoid(BN(Conv_{1\times1}(ReLU(\cdot))))
    \label{eqn:So}
\end{equation}
\begin{equation}
    AG(x_s, x_u) = SO_{am}(GC(x_u) \oplus GC(x_s)) \otimes x_u
    \label{eqn:AM}
\end{equation}

Where, $\oplus$ represents element wise addition and $\otimes$ represents element wise multiplication.

\subsection{Mutation}
\label{sec:mutation}
The segmentation maps $I$ presented in Equation \ref{eqn:outputset} are used for mutation and model optimization.
\par\textbf{{Loss aggregation:}} We employ a combinatorial strategy for loss combination known as \textit{mutation} \cite{Rahman_2024_CVPR}. Unlike EMCAD \cite{Rahman_2024_CVPR}, outputs from both decoders (\textit{noise-free} ($j=1$) and \textit{noisy} ($j=2$)) are mutated. The mutated features combine both \textit{noise-free} and \textit{noisy} properties, thus improving robustness. This method calculates the loss for every possible non-empty subset of predictions produced by the four segmentation logits $I$ (see Equation \ref{eqn:outputset}), resulting in a total of $|\mathcal{P}(I)| = 2^4 - 1 = 15$ unique predictions as defined in Equation \ref{eqn:powerset}. The losses from these combinations are then summed together. As presented in Equation \ref{eqn:loss}, the loss function $\mathcal{L}(\cdot)$ combines Dice loss $\mathcal{L}_{dice}(\cdot)$ and CrossEntropy loss $\mathcal{L}_{ce}(\cdot)$ with weights $\alpha = 0.7$ and $\beta = 0.3$. This training process minimizes the cumulative combinatorial loss $\mathcal{L}_{\text{total}}$ defined in Equation \ref{eqn:all_loss}.
\begin{equation}
    \mathcal{P}(I) = \{A \mid A \subseteq I, A \neq \emptyset\}
    \label{eqn:powerset}
\end{equation}
\begin{equation}
\label{eqn:loss}
    \mathcal{L}(\cdot) = \alpha \mathcal{L}_{dice}(\cdot) + \beta \mathcal{L}_{ce}(\cdot)
\end{equation}
\begin{equation}
\label{eqn:all_loss}
    \mathcal{L}_{\text{total}} = \sum_{i=1}^{15} \mathcal{L}\left(\text{softmax}\left(\sum_{x \in P_i} x\right), \text{target} \right)
\end{equation}

\subsection{Final Segmentation Mask Generation:}
For inference or real-time applications, only the prediction mask $I_1^1$ is utilized as the final output where only the encoder and the noise-free decoder ($j = 1$) is necessary. Although the noisy decoder ($j = 2$) is employed during training to guide the noise-free decoder towards improved robustness and noise adaptability, it is unnecessary during testing, deployment or real-time applications. The final segmentation results are generated by applying the \textit{Softmax} function for multi-class segmentation, as described in equation \ref{eqn:output}.
\begin{equation}
\label{eqn:output}
    Output = Softmax(I_1^1)
\end{equation}

\begin{table*}[ht]
\centering
\caption{The segmentation performance of abdominal organs on the Synapse Multi-organ dataset is evaluated. Dice scores reported for each organ individually. $\uparrow$ and $\downarrow$ denote the higher the better and lower the better, respectively. The MACs count for all methods are reported for input sizes of $256 \times 256$. A dash (`-') indicates missing data from the source, and the best results are highlighted in \textbf{bold}. }
\begin{adjustbox}{width=\textwidth}
\begin{tabular}{l|r|c|c|c|c|c|c|c|c|c|c}
 \hline
\multirow{2}{*}{\textbf{Method}} & \multirow{2}{*}{\textbf{MACs $\downarrow$}}& \multirow{2}{*}{\textbf{DICE \% $\uparrow$}} & \multirow{2}{*}{\textbf{mIoU \% $\uparrow$}} & \multicolumn{8}{c}{\textbf{Organ (Dice \% $\uparrow$)}} \\
 \cline{5-12}
 
& & & &\textbf{Aorta} &\textbf{GB} & \textbf{KL} & \textbf{KR} & \textbf{Liver} & \textbf{PC} & \textbf{SP} & \textbf{SM} \\
\hline

UNet \cite{Ronneberger2015} & 54.77G & 70.11 & 59.39 & 84.00 & 56.70 & 72.41 & 62.64 & 86.98 & 48.73 & 81.48 & 67.96 \\ \hline
AttnUNet \cite{Ozan2018} & 66.67G & 71.70 & 61.38 & 82.61 & 61.94 & 76.07 & 70.42 & 87.54 & 46.70 & 80.67 & 67.66 \\ \hline
R50+UNet \cite{chen2021transunet} & 85.11G & 74.68 & -- & 84.18 & 64.82 & 79.19 & 71.29 & 93.35 & 45.23 & 84.41 & 73.92 \\ \hline
R50+AttnUNet \cite{chen2021transunet} & 91.72G & 75.57 & -- & 85.92 & 63.91 & 79.20 & 72.71 & 93.78 & 45.19 & 84.79 & 74.95 \\ \hline
SSFormer \cite{10.1007/978-3-031-16437-8_11} & 17.28G & 78.01 & 67.23 & 86.50 & 70.22 & 78.11 & 73.78 & 93.53 & 61.53 & 87.07 & 76.61 \\ \hline
PolypPVT \cite{dong2023PolypPVT} & 5.3G & 78.08 & 67.43 & 88.05 & 66.14 & 81.21 & 73.87 & 94.33 & 59.34 & 87.00 & 79.40 \\ \hline
TransFuse \cite{transFUse} & 82.71G & 77.42 & - & 85.15 & 63.06 & 80.57 & 78.58 & 94.22 & 57.06 & 87.03 & 73.69 \\ \hline
TransUNet \cite{chen2021transunet} & 32.46G & 77.61 & 67.32 & 86.56 & 63.43 & 78.11 & 73.85 & 94.37 & 58.47 & 86.84 & 75.00 \\ \hline
DS-TransUNet \cite{ds-transunet}  & 51.02G & 78.13 & - & 86.11 & 63.59 & 83.63 & 78.72 & 94.36 & 57.26 & 87.88 & 73.50 \\ \hline
SwinUNet \cite{cao2021swinunet} & 6.2G & 77.58 & 66.88 & 81.76 & 65.95 & 82.32 & 77.22 & 94.35 & 53.81 & 86.84 & 75.79 \\ \hline
MT-UNet \cite{wu2021mtunet} & 51.97G & 78.96 & - & 88.55 & 68.65 & 82.10 & 77.29 & 94.41 & 65.67 & 91.92 & 80.81 \\ \hline
UDBANet \cite{10635587} & 24.59G & 79.99 & 70.02 & 88.73 & 66.50 & 87.07 & 81.99 & 94.62 & 57.86 & 87.80 & 74.53 \\ \hline
PVT-CASCADE \cite{Rahman_2023_WACV} & 8.12G & 81.06 & 70.88 & 83.01 & 70.59 & 82.23 & 80.37 & 94.08 & 64.43 & 91.31 & 83.52 \\ \hline
TransCASCADE \cite{Rahman_2023_WACV} & 22.47G & 82.68 & 73.48 & 88.48 & 68.48 & 87.66 & \textbf{84.56} & 94.45 & 65.33 & 90.79 & 83.52 \\ \hline
PVT-GCASCADE-B2 \cite{rahman2024g} & 5.8G & 83.28 & 73.91 & 86.50 & 71.71 & 87.07 & 83.77 & 95.31 & 66.72 & 90.84 & 83.58 \\ \hline
PVT-EMCAD-B2 \cite{Rahman_2024_CVPR} & 5.6G & 83.63 & 74.65 & 88.14 & 68.87 & \textbf{88.08} & 84.10 & 95.26 & \textbf{68.51} & \textbf{92.17} & 83.92 \\ \hline
EDLDNet (Our) & 5.6G & \textbf{84.00} & \textbf{75.03} & \textbf{89.12} & \textbf{74.15} & 87.27 & 82.63 & \textbf{95.56} & 67.89 & 91.34 & \textbf{83.99} \\ \hline

\end{tabular}
\end{adjustbox}
\label{tab:synapse}
\end{table*}

\section{Experiments and Results}
\label{sec:exp and result}

\subsection{Dataset}
To evaluate the effectiveness of the proposed method, we use four publicly available multi-organ segmentation datasets including \emph{Synapse} \cite{synapse_dataset} \footnote{\url{https://doi.org/10.7303/syn3193805}}, \emph{ACDC} \footnote{\url{https://www.creatis.insa-lyon.fr/Challenge/acdc/index.html}} \cite{8360453}, \emph{SegThor} \footnote{\url{https://competitions.codalab.org/competitions/21145}} \cite{9286453}, and \emph{LCTSC} \footnote{\url{https://doi.org/10.7937/K9/TCIA.2017.3R3FVZ08}} \cite{LCTSC}.
\subsubsection{Synapse}
It comprises 30 cases, in total 3,779 axial abdominal CT images, annotated for the segmentation of eight organs: aorta, gallbladder, spleen, left kidney, right kidney, liver, pancreas, and stomach. Each CT volume consists of 85 $\sim$ 198 slices of $512 \times 512$ pixels, with a voxel spatial resolution of 
$([0.54 \sim 0.54] \times [0.98 \sim 0.98] \times [2.5 \sim 5.0]) \, \text{mm}^3$. Consistent with the setup in TransUNet \cite{chen2021transunet}, the dataset is divided into 18 cases for training and 12 cases for testing.

\subsubsection{ACDC}
The ACDC challenge involves the collection of dataset from various patients using MRI scanners. The slices have a thickness ranging from 5 mm to 8 mm, and the spatial resolution of the short-axis in-plane varies between 0.83 mm²/pixel and 1.75 mm²/pixel \cite{8360453}. Each patient’s scan is annotated manually to provide ground truth labels for the left ventricle (LV), right ventricle (RV), and myocardium (MYO). Consistent with the setup in TransUNet \cite{chen2021transunet} 70 cases (comprising 1930 axial slices) are used for training, 10 cases for validation, and 20 cases for testing.

\subsubsection{SegThor}  
The SegThor dataset consists of CT scans from 40 patients, with manual annotations for four organs at risk: Esophagus, Heart, Trachea, and Aorta. For model training, data from 32 patients are used, while the remaining 8 patients' data are reserved for testing. The dataset includes a total of 7,390 slices, each with a resolution of $512 \times 512$ pixels \cite{9286453}.

\subsubsection{LCTSC}  
The LCTSC dataset contains CT scans and corresponding annotations for five organs: Esophagus, Spinal Cord, Heart, Left Lung, and Right Lung. It consists of data from 60 patients, with 36 patients' data used for training, 12 patients' data for testing, and 12 patients' data for validation. The dataset comprises 9,593 slices, each with a resolution of $512 \times 512$ pixels \cite{LCTSC}.

\subsection{Experimental Parameters}
To conduct experiments, we used 300 epoch for synapse, and ACDC and 200 epoch for SegThor, and LCTSC in train and the batch size was 6. For optimization, AdamW optimizer are used  with learning rate 0.0001 and weight decay 0.0001. The input size is $224 \times 224$ for Synapse, and ACDC dataset and $256 \times 256$ for SegThor, and LCTSC dataset. To evaluate the performance of different segmentation methods, we utilize popular metrics Dice score, mean Intersection over Union (mIoU), Asymmetric Surface Distance (ASD) \cite{jia2024segmetricspythonpackagecompute}. For evaluating the required computation of a segmentation method, we use Multiply-Accumulate Operations (MACs) count \cite{getzner2023accuracy}.

\subsection{Environment}
\label{sec:environment}
The experimental setup utilizes Python 3.8 with PyTorch version 1.11.0+cu113 for implementation. Model training and evaluation are conducted on a high-performance computing system equipped with an Intel Xeon processor running at 2.40 GHz, 64 GB of RAM, and an Nvidia V100 GPU.

\begin{table}[ht]
\centering
\caption{The segmentation results for cardiac organs on the ACDC dataset are presented, with Dice scores (\%) reported for each organ individually. $\uparrow$ denotes the higher the better. The best results are highlighted in bold.}
\begin{adjustbox}{width=0.47\textwidth}
\begin{tabular}{{l}|{c}|{c}|{c}|{c}}
\hline
\multirow{2}{*}{\textbf{Method}} & \multirow{2}{*}{\textbf{DICE \% $\uparrow$}} & \multicolumn{3}{c}{\textbf{Organ (Dice \% $\uparrow$)}} \\ 
\cline{3-5}

& & \textbf{RV} & \textbf{Myo} & \textbf{LV} \\

\hline
R50+UNet \cite{chen2021transunet} & 87.55 & 87.10 & 80.63 & 94.92 \\ \hline
R50+AttnUNet \cite{chen2021transunet} & 86.75 & 85.78 & 79.20 & 93.47 \\ \hline
ViT+CUP \cite{chen2021transunet} & 81.45 & 81.46 & 70.71 & 92.18 \\ \hline
R50+ViT+CUP \cite{chen2021transunet} & 87.57 & 87.11 & 80.79 & 94.95 \\ \hline
TransUNet \cite{chen2021transunet} & 89.71 & 86.76 & 87.19 & 95.60 \\ \hline
SwinUNet \cite{cao2021swinunet} & 88.07 & 85.77 & 84.42 & 94.03 \\ \hline
MT-UNet \cite{wu2021mtunet} & 90.43 & 86.64 & 89.04 & 95.62 \\ \hline
MISSFormer \cite{huang2021missformer} & 90.46 & 86.94 & 88.04 & 95.70 \\ \hline
UDBANet \cite{10635587} & 90.41 & 87.73 & 88.30 & 92.20 \\ \hline
PVT-CASCADE \cite{Rahman_2023_WACV} & 91.46 & 89.97 & 88.99 & 95.50 \\ \hline
TransCASCADE \cite{Rahman_2023_WACV} & 91.63 & 90.25 & 89.14 & 95.50 \\ \hline
PVT-GCASCADE-B2 \cite{rahman2024g} & 91.95 & 90.31 & 89.63 & 95.91 \\ \hline
PVT-EMCAD-B2 \cite{Rahman_2024_CVPR} & 92.12 & 90.65 & 89.68 & \textbf{96.02} \\ \hline
EDLDNet (Our) & \textbf{92.25} & \textbf{91.20} & \textbf{89.75} & 95.82 \\ \hline

\end{tabular}
\end{adjustbox}
\label{tab:comparison}
\end{table}

\begin{table*}[ht]	
            \caption{Comparison between other baselines on SegThor Dataset. The results of AttUnet, UNet++, R2UNet and UDBANet taken form \cite{10635587}. $\uparrow$ and $\downarrow$ denote the higher the better and lower the better, respectively. The best results are highlighted in bold.}
		\label{Tab:SOTA_Segthor}
		\centering
    \begin{adjustbox}{width=\textwidth}
		\begin{tabular}{l |{c}| {c}| {c}| {c}|{c}|{c}| {c}| {c} |{c}| {c}| {c}| {c}}
			\hline
			\multirow{2}{*}{\textbf{Method}}&\multicolumn{3}{|c|}{\textbf{ Esophagus }} &\multicolumn{3}{c|}{\textbf{ Heart }} &\multicolumn{3}{c|}{\textbf{ Trachea }} & \multicolumn{3}{c}{\textbf{ Aorta}}\\
		    \cline{2-13} &\textbf{Dice$\uparrow$}&\textbf{ASD$\downarrow$}&\textbf{mIoU$\uparrow$} &\textbf{Dice$\uparrow$}&\textbf{ASD$\downarrow$}&\textbf{mIoU$\uparrow$} &\textbf{Dice$\uparrow$}&\textbf{ASD$\downarrow$}&\textbf{mIoU$\uparrow$} &\textbf{Dice$\uparrow$}&\textbf{ASD$\downarrow$}&\textbf{mIoU$\uparrow$} \\
             \hline
            AttUnet \cite{Ozan2018}     &0.74&\textbf{0.66}&0.89 &0.80&1.57&0.61 &0.80&0.50&0.86 &0.88&0.59&0.88\\
            \hline
            UNet++ \cite{zhou2020unet}   &0.81&0.82&0.93	&0.95&1.00&0.89	 &0.89&0.46&0.93	&0.93&0.55&0.91 \\
			\hline
            R2UNet \cite{8556686}  &0.69&0.86&0.83 	&0.77&1.97&0.63	 &0.81&0.50&0.83	 &0.81&0.90&0.83\\
            \hline
            UDBANet \cite{10635587}      &0.81&1.45&0.89   &0.95&0.94&0.93  
            &0.91&0.46&0.95  &0.93&0.66&0.93\\
            \hline
            PVT-CASCADE \cite{Rahman_2023_WACV}  &   0.81&0.75&0.93  &0.95&0.77&0.93  &0.91&0.43&0.95  &0.94&\textbf{0.50}&\textbf{0.95}\\
            \hline
            PVT-GCASCADE-B2 \cite{rahman2024g} & 0.81 & 0.78 & 0.93 & 0.95 & 0.82 & 0.93 & 0.90 & 0.44 & 0.95 & 0.94 & 0.54 & 0.94 \\ \hline
            PVT-EMCAD-B2 \cite{Rahman_2024_CVPR}      &0.81&0.75&0.93   &0.95&\textbf{0.73}&\textbf{0.94}  &\textbf{0.92}&0.43&\textbf{0.96}  &0.94&0.51&0.94\\
            \hline
            
           EDLDNet (Our)   &\textbf{0.82}&0.74&\textbf{0.94}   &\textbf{0.96}&0.79&\textbf{0.94}  &\textbf{0.92}&\textbf{0.42}&\textbf{0.96}&\textbf{0.95}&\textbf{0.50}&\textbf{0.95}\\
            \hline
		\end{tabular}
  \end{adjustbox}
\end{table*}

\begin{table*}[ht]
		\caption{Comparison between other baselines on LCTSC Dataset. The results of AttUnet, UNet++, R2UNet and UDBANet taken form \cite{10635587}. $\uparrow$ and $\downarrow$ denote the higher the better and lower the better, respectively. The best results are highlighted in bold.}
		\label{Tab:SOTA_LCTSC}
		\centering
        \begin{adjustbox}{width=\textwidth}
		\begin{tabular}{l |{c}| {c}| {c}| {c}|{c}|{c}| {c}| {c} |{c}| {c}| {c}| {c}| {c}| {c}| {c}}
			\hline
			\multirow{2}{*}{\textbf{Method}} &\multicolumn{3}{|c|}{\textbf{Esophagus}} &\multicolumn{3}{c|}{\textbf{Spine}} &\multicolumn{3}{c|}{\textbf{Heart}} &\multicolumn{3}{c}{\textbf{Lung(L)}} &\multicolumn{3}{|c}{\textbf{Lung(R)}}\\
		    \cline{2-16}
             &\textbf{Dice$\uparrow$}&\textbf{ASD$\downarrow$}&\textbf{mIoU$\uparrow$} &\textbf{Dice$\uparrow$}&\textbf{ASD$\downarrow$}&\textbf{mIoU$\uparrow$} &\textbf{Dice$\uparrow$}&\textbf{ASD$\downarrow$}&\textbf{mIoU$\uparrow$} &\textbf{Dice$\uparrow$}&\textbf{ASD$\downarrow$}&\textbf{mIoU$\uparrow$} &\textbf{Dice$\uparrow$}&\textbf{ASD$\downarrow$}&\textbf{mIoU$\uparrow$}\\
             \hline
            AttUnet \cite{Ozan2018}     &0.66&\textbf{1.09}&0.70     &\textbf{0.89}&\textbf{0.66}&0.91   &0.69&3.26&0.39   &0.95&0.68&0.86   &0.96&0.72&0.90\\
            \hline
            UNet++ \cite{zhou2020unet}  &0.72&1.14&0.71	&\textbf{0.89}&\textbf{0.66}&0.91	 &0.89&1.63&0.51	&0.96&0.69&0.87	&0.96&0.73&0.89\\  
			\hline
        
            R2UNet \cite{8556686} &0.51&1.74&0.54	&0.62&0.95&0.89	 &0.55&3.34&0.31	    &0.94&0.73&0.78	    &0.95&0.76&0.78\\
            \hline
            UDBANet \cite{10635587}      &0.71&1.58&0.80&\textbf{0.89}&\textbf{0.66}&0.91  &0.92&1.35&0.84 &\textbf{0.97}&0.62&0.92
            &\textbf{0.97}&0.68&0.92\\
            \hline
            
            PVT-CASCADE \cite{Rahman_2023_WACV}     &\textbf{0.74}&1.19&0.86
            &0.88&0.67&0.91  
            &0.93&\textbf{1.11}&\textbf{0.89} &0.96&0.60&0.92  &0.96&0.65&0.92\\
            \hline

            PVT-GCASCADE-B2 \cite{rahman2024g} & \textbf{0.74} & 1.15 & 0.86 & 0.88 & 0.67 & 0.91 & 0.93 & \textbf{1.11} & 0.88 & 0.96 & 0.57 & 0.92 & 0.96 & \textbf{0.64} & 0.92 \\ \hline

            PVT-EMCAD-B2 \cite{Rahman_2024_CVPR}      &\textbf{0.74}&1.20&0.86 
            &0.88&\textbf{0.66}&0.91  
            &0.93&\textbf{1.11}&0.88&0.96&\textbf{0.56}&0.92  &0.96&\textbf{0.64}&0.92\\
            \hline

           EDLDNet (Our)&\textbf{0.74}&1.20&\textbf{0.87} 
            &\textbf{0.89}&\textbf{0.66}&\textbf{0.92}  
            &\textbf{0.94}&\textbf{1.11}&\textbf{0.89} &\textbf{0.97}&0.59&\textbf{0.93}  &\textbf{0.97}&\textbf{0.64}&\textbf{0.93}\\
            \hline
		\end{tabular}
  \end{adjustbox}
\end{table*}

\begin{figure*}[!htb]
\includegraphics[width=\linewidth, trim={1.5cm 15.1cm 1.5cm 0}, clip]{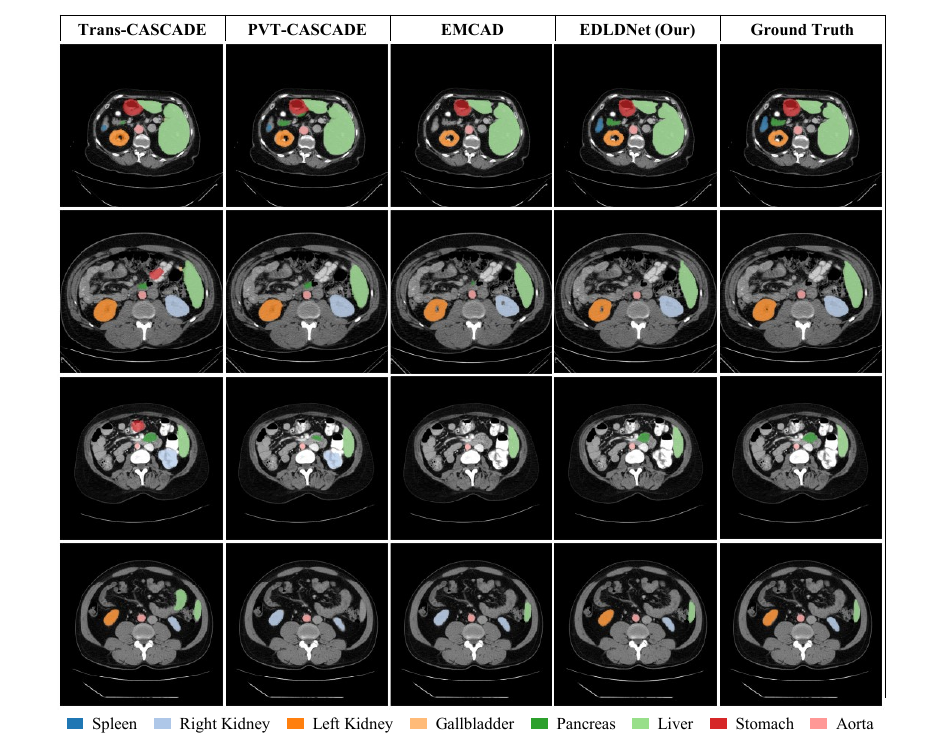}

\caption{The comparison of contoured segmentation images from Synapse dataset among proposed method and the competitive existing methods.} 
\label{fig:synapse_contoured}
\end{figure*}

\subsection{Results on Synapse Dataset}  
Table \ref{tab:synapse} presents a detailed quantitative comparison and Figure \ref{fig:synapse_contoured} presents the quantitative comparison of various segmentation models evaluated on the Synapse Multi-Organ Dataset. The performance of these models is measured using Dice score, mIoU, and computational efficiency in terms of MACs. Given the complexity of the dataset—comprising multiple abdominal organs of varying shapes, sizes, and textures—achieving high segmentation accuracy is a significant challenge. The results show a clear evolution from traditional CNN-based models to Transformer-based architectures, with the latter demonstrating remarkable improvements in segmentation accuracy while also optimizing computational efficiency.

Among the earlier models, UNet serves as a baseline, achieving a Dice score of 70.11\%. Its improved version, Attention UNet, incorporates attention mechanisms to focus on relevant spatial regions, slightly boosting performance to 71.70\%. However, these models struggle with segmenting smaller and more complex organs like the pancreas and gallbladder. The introduction of hybrid architectures such as R50+UNet and R50+Attention UNet further enhances performance, pushing Dice scores to 74.68\% and 75.57\%, respectively. Despite these improvements, the increased computational cost is a major drawback—R50+Attention UNet requires 91.72G MACs, making it impractical for real-time applications. Transformer-based models have significantly transformed medical image segmentation by leveraging self-attention mechanisms to capture long-range dependencies. TransUNet, one of the first Transformer-based models in this domain, achieves a notable jump in performance with a Dice score of 77.61\%, while being computationally more efficient than its ResNet-based UNet counterparts (32.46G MACs). More recent Transformer-based architectures, such as SSFormer, PolypPVT, and MT-UNet, continue this trend of improvement. Notably, MT-UNet achieves a Dice score of 78.96\%, while PolypPVT delivers a similar 78.08\% with lower computational demands. UDBANet utilizes uncertainty information within the network and it achieves 79.99\% dice score with 24.59G MACs. Among these, PVT-CASCADE stands out, reaching an impressive Dice score of 81.06\% while keeping MACs at just 8.12G, highlighting the power of PVT in achieving both high accuracy and efficiency. At the top end of the performance spectrum, GCASCADE, EMCAD, and the proposed model achieve SOTA segmentation results while maintaining efficiency. GCASCADE delivers a Dice score of 83.28\%, while EMCAD refines it further to 83.63\%, with an improved mIoU of 74.65\%. The proposed model, EDLDNet, achieves SOTA performance with an 84.00\% Dice score and 75.03\% mIoU, surpassing all previous methods. Despite this significant performance improvement, the model remains highly computationally efficient, requiring only 5.6G MACs, making it well-suited for real-world applications. EDLDNet outperforms baseline methods like UNet by 20\% in accuracy while reducing computational cost by 89\% in terms of MACs Operations. Compared to recent approaches such as EMCAD, EDLDNet demonstrates marginal yet consistent improvements in segmentation accuracy while maintaining comparable efficiency, underscoring its ability to balance high performance with low computational complexity.

At the organ level performance, the proposed model achieves the highest Dice scores for the aorta (89.12\%), gallbladder (74.15\%), liver (95.56\%), and stomach (83.99\%), demonstrating its robustness in segmenting both large and small organs. While EMCAD performs slightly better on the pancreas (68.51\%) and spleen (92.17\%), our model remains highly competitive, achieving 67.89\% and 91.34\%, respectively. Kidney segmentation remains particularly challenging, with EMCAD achieving the best Dice score for the left kidney (88.08\%), closely followed by proposed model at 87.27\%. For the right kidney, TransCASCADE edges out our model (84.56\% vs. 82.63\%), indicating areas where further refinements could improve performance. This model strikes an ideal balance, achieving SOTA accuracy while maintaining an extremely low computational requirement as presented in Figure \ref{fig:mac_vs_dice} which makes it a viable solution for practical applications where both precision and efficiency are critical.

\begin{figure*}[!htb]
\includegraphics[width=\linewidth, trim={1.5cm 15.1cm 1.5cm 0}, clip]{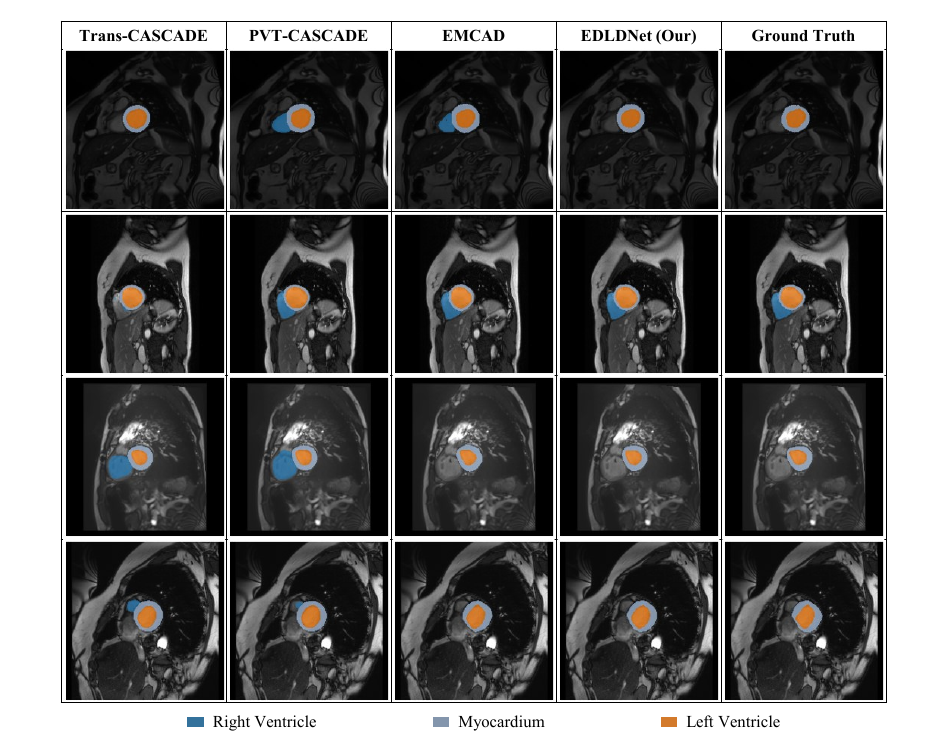}

\caption{The comparison of contoured segmentation images from ACDC dataset among proposed method and the competitive existing methods.} 
\label{fig:ACDC_contoured}
\end{figure*}

\subsection{Results on ACDC Dataset}
The quantitative comparative analysis in Table \ref{tab:comparison} and qualitative analysis in Figure \ref{fig:ACDC_contoured} highlight the evolution of cardiac segmentation methods, emphasizing the performance of SOTA techniques across three anatomical structures: the Right Ventricle (RV), Myocardium (Myo), and Left Ventricle (LV). Traditional convolutional approaches, such as R50+UNet and R50+AttnUNet, show moderate overall Dice scores of 87.55\% and 86.75\%, respectively, due to their limitations in capturing global contextual information. Transformer-based models like TransUNet and SwinUNet significantly improved segmentation accuracy, with TransUNet achieving an overall Dice score of 89.71\% by leveraging long-range dependency modeling. While uncertainty based segmentation method UDBANet achieves 90.41\%,  the cascaded architectures, including PVT-CASCADE, TransCASCADE, and  GCASCADE further advanced performance with overall Dice scores of 91.46\%, 91.63\%, and 91.95\%.

Our proposed method sets a new benchmark with the highest overall Dice score of 92.25\%, outperforming all prior methods, particularly in RV and Myo segmentation. While EMCAD slightly surpasses in LV segmentation, the proposed method demonstrates exceptional generalization across all structures, underscoring its robustness. It also reveals the persistent challenge of Myocardium segmentation, which lags behind RV and LV due to its indistinct boundaries and structural variability.

\subsection{Result on SegThor Dataset}
The comparison in Table \ref{Tab:SOTA_Segthor} evaluates segmentation methods for four anatomical structures—Esophagus, Heart, Trachea, and Aorta—using metrics like Dice similarity, ASD, and mIoU. AttUNet performs well in ASD for the Esophagus but falls behind in Dice and mIoU, which indicates challenges with boundary segmentation accuracy. Similarly, R2UNet struggles across all metrics, especially for the Esophagus and Heart, reflecting its limitations in capturing complex anatomical details. Advanced methods like UNet++ and UDBANet show notable improvements, with UNet++ excelling in Heart and Trachea segmentation (Dice of 0.95 and 0.89) and UDBANet performing strongly for the Heart and Aorta, though its precision for the Esophagus is limited by a high ASD of 1.45.

Recent methods such as PVT-CASCADE, PVT-GCASCADE-B2, and PVT-EMCAD-B2 demonstrate competitive performance, with PVT-CASCADE achieving strong results in Dice and mIoU for all organs and the lowest ASD for the Aorta. PVT-EMCAD-B2 further improves upon these results, particularly in Heart and Trachea segmentation, achieving the highest mIoU and competitive ASD values. However, the proposed EDLDNet method surpasses all other approaches, achieving the highest Dice and mIoU scores for all organs and the lowest ASD values for the Heart, Trachea, and Aorta. For challenging regions like the Esophagus, it records significant improvements despite inherent segmentation difficulties due to its variable shape. 

\subsection{Results on LCTSC Dataset}
The results in Table \ref{Tab:SOTA_LCTSC} compare segmentation performance across eight methods for the Esophagus, Spine, Heart, and Left and Right Lungs (Lung(L) and Lung(R)), evaluated using Dice score, ASD, and mIoU. For the Esophagus, the proposed method (EDLDNet) achieves the highest Dice (0.74) and mIoU (0.87), significantly outperforming others, while AttUNet records the lowest ASD (1.09) but with lower Dice (0.66) and mIoU (0.70), indicating weaker region segmentation. Spine segmentation shows consistent performance across methods, with Dice values of 0.88-0.89 and mIoU of 0.91-0.92. AttUNet, UNet++, UDBANet, and the proposed method achieve the lowest ASD (0.66), suggesting that the Spine is relatively easier to segment.

Recent methods such as PVT-CASCADE, PVT-GCASCADE-B2, and PVT-EMCAD-B2 demonstrate competitive performance across all organs. PVT-CASCADE achieves strong results in Heart segmentation, with a Dice score of 0.93 and the lowest ASD of 1.11, while PVT-EMCAD-B2 matches these results and further improves ASD for the Lungs. PVT-GCASCADE-B2 also performs well, achieving the lowest ASD for Lung(R) (0.64). However, the proposed method consistently outperforms these approaches, achieving the highest Dice (0.94) and mIoU (0.89) for the Heart, along with the lowest ASD (1.11). For the Lungs, the proposed method demonstrates exceptional performance, with the highest Dice (0.97) and mIoU (0.93) for both lungs and competitive ASD values (0.59 for Lung(L) and 0.64 for Lung(R)). While UDBANet matches our Dice scores for the lungs, its slightly higher ASD values reflect less precise boundary alignment. Other methods, such as AttUNet and UNet++, show competitive but less precise results, and R2UNet consistently underperforms. Overall, the proposed method achieves state-of-the-art performance across all metrics and organs, demonstrating its robustness and precision in medical image segmentation.

\subsection{Ablation Studies}
\begin{table}[ht]
\centering
\caption{Performance comparison of our method with different configurations. $\uparrow$ denotes the higher the better. The best results are highlighted in bold.}

\begin{tabular}{c|c|c|c|c}
\hline
\multicolumn{4}{c|}{\textbf{Components}} & \multirow{2}{*}{DICE \% $\uparrow$} \\
\cline{1-4}
UCB & AG & MSCAM & Noise  &           \\
\hline
$\surd$ & - & - & - & 81.08 \\ \hline
$\surd$ & $\surd$ & - & - & 81.92 \\ \hline
$\surd$ & - & $\surd$ & - & 82.86 \\ \hline
$\surd$ & $\surd$ & $\surd$ & - &  83.60 \\ \hline
$\surd$ & $\surd$ & $\surd$ & $\surd$ & \textbf{84.00} \\ \hline
\end{tabular}
\label{tab:ablation}
\end{table}
We performed an ablation study by adding various elements to the baseline model which is presented in. Table \ref{tab:ablation}. It indicates that the feature extraction achieved with only UCB helps to reach DICE score of 81.08\%. The DICE score rises to 81.92\% when incorporating the AG, suggesting that AG enhances feature selection by highlighting pertinent areas while mitigating background noise. The further integration of MSCAM into the model achieves an even higher DICE score of 82.86\%, indicating that multi-scale attention fosters learning of contextual feature information, improving segmentation accuracy. Nonetheless, the best improvement is seen when structural noise is introduced in the second decoder $(j=2)$, resulting in the maximum DICE score of 84.00\% recorded. The gain indicates that addition of noise operates as an implicit regularizer that improves feature generalization and reduces overfitting to provide robustness. Adding noise drives the network to learn more discriminative and invariant representations. The performance gains are consistent and demonstrate that such controlled perturbations in the training process enable the model to handle real-world shifts and out-of-distribution environments better.  

The addition of noise acts as an implicit regularizer, significantly enhancing segmentation performance. Noise-driven learning provides greater generalization capability by forcing the model to correlate different manifestations or variations of the same information, such that this approach has particular utility in medical imaging applications where variations due to real-world circumstances are a regular occurrence.

\section{Conclusion}
\label{sec:Conclusion}
In this study, we introduce a new segmentation network termed as EDLDNet for accurate and efficient organ-at-risk segmentation. Our approach integrates multi-scale convolutional attention and noise regularization to enhance performance while keeping computational costs low. Unlike traditional segmentation networks, we use a noisy decoder during training to improve robustness, but for inference, we retain only the noise-free decoder. This strategy improves segmentation accuracy without increasing computation cost of inference. We implement a mutation-based loss function that combines multiple multi-scale segmentation outputs from both decoders to improve generalization. To evaluate its effectiveness, we tested our model on four publicly available datasets—Synapse, ACDC, SegThor, and LCTSC. The results show that our approach outperforms SOTA methods in both accuracy and efficiency. Specifically, our model achieves an 84\% Dice score on the Synapse dataset which reflects a 13.89\% improvement over baseline methods, while also reducing computational costs by 89.7\% in terms of MACs count. These results highlight the model’s ability to handle diverse anatomical structures with high precision that makes it a strong candidate for real-world applications.  

Looking ahead, we plan to extend our approach to integrate uncertainty-aware segmentation to provide confidence estimates, which could support clinical decision-making. Future research will explore self-supervised learning (SSL) and lightweight deployment on edge AI devices, enabling real-time, low-resource applications. These advancements will further strengthen the role of our method in precision medicine and automated diagnostics, positioning it as a cutting-edge solution in medical image segmentation.

\section*{Acknowledgment}
This work was partially supported by the Innovation Fund from ICT Division, Bangladesh under Reference No. 56.00.0000.053.33.0002.24.418.

 \bibliographystyle{elsarticle-num} 
 \bibliography{reference}
\end{document}